\documentclass{article}

\usepackage[final]{nips_2017}
\usepackage[utf8]{inputenc} 
\usepackage[T1]{fontenc}    
\usepackage{hyperref}       
\usepackage{url}            
\usepackage{booktabs}       
\usepackage{amsfonts}       
\usepackage{microtype}      

\usepackage{amsmath}
\usepackage{amsfonts}
\usepackage{natbib}
\usepackage{bm}
\usepackage{amsthm}
\usepackage[mathscr]{euscript}
\usepackage{nicefrac}
\usepackage{drawnn}
\usepackage{caption}
\usepackage{algorithm, algpseudocode}

\newcommand{\ignore}[1]{}

\newcommand{\set}[2][]{#1 \{ #2 #1 \} }

\newtheorem*{rep@theorem}{\rep@title}
\newcommand{\newreptheorem}[2]{%
\newenvironment{rep#1}[1]{%
 \def\rep@title{#2 \ref{##1}}%
 \begin{rep@theorem}}%
 {\end{rep@theorem}}}

\newtheorem{theorem}{Theorem}
\newreptheorem{theorem}{Theorem}

\newreptheorem{lemma}{Lemma}
\newtheorem{definition}[theorem]{Definition}

\newreptheorem{corollary}{Corollary}

\newreptheorem{proposition}{Proposition}

\title{EnergyNet: Energy-based Adaptive Structural Learning of
  Artificial Neural Network Architectures}

\author{
  Gus Kristiansen\thanks{Work done as a member of the Google Brain Residency
    program (\url{g.co/brainresidency}.)} \\
  Google Research \\
  New York, NY 10011 \\
  \texttt{\small gusatb@google.com} \\
  \And
  Xavi Gonzalvo \\
  Google Research \\
  New York, NY 10011 \\
  \texttt{\small xavigonzalvo@google.com}
}

\begin{document}

\maketitle

\begin{abstract}
We present \textsc{EnergyNet}, a new framework for analyzing and
building artificial neural network architectures. Our approach
adaptively learns the structure of the networks in an unsupervised
manner. The methodology is based upon the theoretical guarantees of
the energy function of restricted Boltzmann machines (RBM) of infinite
number of nodes. We present experimental results to show that the
final network adapts to the complexity of a given problem.
\end{abstract}


\section{Introduction}
\label{sec:intro}
Despite the huge success of deep learning during the last years as a
powerful framework, choosing the right architecture still remains a
practical problem.
%
%
This presents several problems. On the one hand, from the point of
view of the optimization problem there is no guarantee of the
optimality of the learning objective~\citep{cortes16adanet_embdnn}. On
the other hand, large-scale hyperparameter tuning requires large
amounts of data and resources (e.g. random
search~\citep{BergstraBardenetBengioKegl2011}).

In this paper, we attempt to alleviate the problem of choosing the
right network architecture.
We interpret the general structure of a neural network (NN) as an
effective generative model for unsupervised learning. We then
introduce \textsc{EnergyNet} as a framework for adapting the structure
and complexity of the network to the difficulty of the particular
problem at hand with no pre-defined architecture.
The adaptation process is divided into two stages. Firstly, the
architecture of the neural network is predicted. Starting from a
simple single layer neural network, we add more neurons as
required.
Once the architecture of the network is estimated it works as a
regular feedforward NN by adding a supervised output layer and
training with gradient descent.

Automatic design of NN architectures has been studied
before. \cite{fahlman89cascade} presented a system with a minimal
network, then automatically trained and added new hidden
units. Genetic algorithms have also been
proposed~\citep{vonk95evolutionary_book, rolls00genetic,
  arifovic01genetic, saemi07genetic, stanley03evolving,
  stanley02augmenting, miikkulainen17evolving}.
Additionally, smaller networks has been used to train larger
networks~\citep{chen15net2net, ha16hyper}. Finally, reinforcement
learning has also been used for searching different
architectures~\citep{zoph16rl}.

Recently~\cite{cortes16adanet} proposed the
\textsc{AdaNet} framework that adaptively learns NN architectures. The
major difference of our work with that framework is as
follows. Firstly, in our case the architecture of the network has no
output layer connections.
Secondly, the architecture of the network is now independent of the
learned weights.

This paper is organized as follows. Section~\ref{sec:preliminaries}
presents the theory of the framework.
We describe the model complexity to balance the size of the networks
and the theoretical properties of the iRBMs. Section~\ref{sec:dbn}
describes the properties of the DBNs. Section~\ref{sec:algo} describes
the algorithm and Section~\ref{sec:experiments} concludes with
experimental results.


\section{Preliminaries and Theory}
\label{sec:preliminaries}
Let $\mathscr{X}$ denote the input space. We assume that training and
test points are drawn i.i.d. according to some distribution
$\mathscr{D}$ over $\mathscr{X} \times \set{0, 1}$. Given any $x \in
\mathscr{X}$, we denote by $\bm{\Phi}(x) \in \mathfrak{R}^{n_0}$, the
feature representation of $x$.


\subsection{Artificial Neural Networks}
\label{sec:theory_nn}
The standard description of a modern feedforward network is a network
of layers of nodes, where each layer is mapped to the layer above it
via a linear mapping composed with a component-wise nonlinear
transformation. To make this description precise, we define a neural
network following~\cite{cortes16adanet_embdnn}. Let $l$ denote the
number of layers in a network. For each $k \in [l]$, denote by $n_k$
the maximum number of nodes in layer $k$.

Let $1 \leq p \leq \infty$ and $k \geq 1$. Then define the set
$\mathscr{H}^{(p)}_{k}$ to be the family of functions at layer $k$ of the
network in the following way:
\begin{align}
  \label{eq:Hk}
  \mathscr{H}^{(p)}_{k} & = \set[\bigg]{x \mapsto \bigg( \sum_{j =
      1}^{n_{k-1}} u_j (\varphi_{k-1}\circ h_j)(x) \bigg)\colon
    \mathbf{u} \in \mathbb{R}^{n_{k-1}}, \| \mathbf{u} \|_p \leq
    \Lambda_k, h_j \in \mathscr{H}^{(p)}_{k - 1}}, \; (\forall k > 1)
\end{align}
where $\mathscr{H}^{(p)}_{1}$ refers to the case of input features,
$\Lambda_k > 0$ is a hyperparameter and where $\varphi_k$ is an
activation function (e.g. Rectified Linear Unit (ReLU),
see~\citep{GoodfellowBengioCourville2016} for more). The choice of
norm $p$ here is left to the learner and will determine both the
sparsity of the network and the accompanying learning guarantee of the
resulting model.

Since neural networks are built as compositions of layers, it is
natural from the theoretical standpoint to first analyze the
complexity of any layer in terms of the complexity of its previous
layer. Results for \textsc{AdaNet}~\citep{cortes16adanet} demonstrate
that this can indeed be done, and that the empirical Rademacher
complexity of any intermediate layer $k$ in the network is bounded by
the empirical Rademacher complexity of its input times a term that
depends on a power of the size of the layer:

\begin{definition}[Complexity of a neural network]
Let $r_\infty = \max_{j \in [1,n_1], i \in[1,m]}
\left|[\mathbf{\Phi}(x_i)]_j\right|$, and $\frac{1}{p} + \frac{1}{q} =
1$. Then for any $k \geq 1$, the empirical Rademacher complexity of
$\mathscr{H}^{(p)}_k$ for a sample $S$ of size $m$ can be upper
bounded as follows:
\begin{equation}
  \label{eq:complexity}
  \widehat{\mathfrak{R}}_S({\cal H}_k^{(p)}) \leq 2^{k - 1} r_\infty
  \Bigg( \prod_{j = 1}^k \Lambda_j n_{j-1}^{\frac{1}{q}} \Bigg)
  \sqrt{\frac{2 \log (2 n_0)}{m}}.
\end{equation}

\end{definition}


\subsection{Restricted Boltzmann Machines (RBM)}

The RBM is a two layer Markov Random Field, where the observed binary
stochastic visible units $\mathbf{v} \in \{0, 1\}^D$ have pairwise
connections to the binary stochastic hidden units $\mathbf{h} \in \{0,
1\}^K$. There are no pairwise connections within the visible units,
nor within the hidden ones~\citep{hinton02rbm}.

In an RBM model, each configuration $(\mathbf{v}, \mathbf{h}) \in
\mathscr{V} \times \mathscr{H}$ has an associated energy value defined by
the following function:

\begin{definition}[RBM energy function]
Restricted Boltzmann machine is an energy-based model, in which we
define the energy for a state $\{\mathbf{v}, \mathbf{h}\}$ as:
\begin{equation}
  \label{eq:rbm_energy}
  E(\mathbf{v}, \mathbf{h}; \bm{\theta}) =
  -\mathbf{v}^\mathrm{T}\mathbf{b}^v
  -\mathbf{h}^\mathrm{T}\mathbf{b}^h -
    \mathbf{v}^\mathrm{T}\mathbf{W}\mathbf{h},
\end{equation}
where $\bm{\theta}=\{\mathbf{W}, \mathbf{b}^v, \mathbf{b}^h \}$ are
the parameters of the model.

\end{definition}

\begin{definition}[Probability distribution of an RBM]
For $m$ data samples, the probability distribution of the RBM of a
visible vector is obtained by marginalizing over all configurations of
hidden vectors:
\begin{equation*}
  \label{eq:rbm_ll}
  P_{\bm{\theta}}(\mathbf{v}) = \frac{1}{Z} \sum_{\mathbf{h}'
    \in \mathscr{H}} \exp(-E(\mathbf{v}, \mathbf{h}'; \bm{\theta}))
\end{equation*}
where $\mathbf{h}_i$ are the hidden nodes of the $i$-th layer, $Z$ is
the partition function used for normalization:
\begin{equation*}
  \label{eq:z}
  Z = \sum_{\mathbf{v}' \in {\cal V}} \sum_{\mathbf{h}' \in {\cal H}}
  \exp(-E(\mathbf{v}, \mathbf{h}; \bm{\theta}))
\end{equation*}

\end{definition}


\subsection{Infinite Restricted Boltzmann Machines (iRBM)}
\label{sec:irbm}
iRBM can be implemented using a random variable~\citep{cote15irbm} or
via Frank-Wolfe optimization~\citep{ping16irbm}.
We follow the former approach where the order of a hidden unit is
taken into account by introducing a random variable $z$ that can be
seen as the effective number of hidden units participating to the
energy. Hidden units are selected starting from the left and the
selection of each hidden unit is associated with an incremental cost
in energy.

The model via the latent variable $z$ allows for different
observations having been generated by a different number of hidden
units. Specifically, for a given $\mathbf{v}$, the conditional
distribution over the corresponding value of $z$ is:
\begin{equation}
  P(z|\mathbf{v}) =
  \frac{N(z)}{\sum_{z'=1}^{K=\infty}N(z')}=\frac{N(z)}{Z(\mathbf{v})}
\end{equation}
where $N(z)=\exp{(-F(\mathbf{v},z))}$ and $K$ is the maximum number of
hidden nodes, which in this case is infinite. The free energy, $F()$
is defined as:
\begin{equation}
  F\left(\mathbf{v},z\right)=-\mathbf{v}^T\mathbf{b}^v-\sum
  _{i=0}^z\left(\text{soft}_+\left(\mathbf{W}_i\mathbf{v}+b_i^h\right)-\beta_i\right)
\end{equation}
where $\beta_i$ is an energy penalty for selecting each $i$-th hidden
unit.

The denominator (i.e. the normalization factor) needs to be
defined for the infinite $K$. In this case $Z(\mathbf{v})$ must be
divided into two parts, before the maximum number of nodes at a
particular time step, $n$, and the rest:
\begin{equation}
  Z(\mathbf{v}) = \sum_{z=1}^{n} N(z) + \sum_{z=n+1}^{\infty} N(z) =
  \sum_{z=1}^{n} N(z)+\frac{a}{a-1}N(z)
\end{equation}
where $a$ is the result of a geometric series of the infinite
term~\citep{cote15irbm}.


\subsection{Deep Belief Networks (DBN)}
\label{sec:dbn}
There are two possible ways of putting together multiple RBMs: Deep
Boltzmann Machines (DBM)~\citep{SalHinton07} and Deep Belief Networks
(DBN)~\citep{hinton06dbn}. Both are probabilistic graphical models
consisting of stacked layers of RBMs. The difference is in how these
layers are connected~\citep{SalHinton07}. In a DBN the connections
between layers are directed. Therefore, the first two layers form an
RBM (an undirected graphical model), then the subsequent layers form a
directed generative model. In a DBM, the connection between all layers
is undirected, thus each pair of layers forms an RBM. If multiple
layers are learned in a greedy, layer-by-layer way, the resulting
composite model is a DBN~\citep{bengio06greedy_layer}.

As described in Section~\ref{sec:algo} the algorithm proposed in this
paper requires the use of the log-likelihood of a DBN. Due to the
presence of the partition function, exact maximum likelihood learning
in RBMs and DBNs are
intractable. In~\citep{salakhutdinov2008quantitative,neil01anneal} an
Annealed Importance Sampling (AIS) is used to efficiently estimate the
partition function of an RBM and to estimate a lower bound on the
log-probability of a DBN. We opted for a more efficient version
presented by~\cite{lucas11dbn}. This still requires an estimation of
the partition function for the top RBM but uses samples drawn from the
densities of each layer in a feed-forward manner.


\section{Algorithm}
\label{sec:algo}
Our System forms a DBN by stacking multiple iRBMs trained layer by
layer. Since an iRBM is an extension of an RBM, the training of a DBN
remains unchanged. After each layer is trained using the contrastive
divergence algorithm applying the specifics of iRBMs (see
Section~\ref{sec:irbm}) it is converted back into a regular RBM in
order to compute the log-likelihood.

The minimum description length (MDL) criterion
(see~\citep{rissanen84universal}) formalizes the idea that for a set
of hypotheses ${\cal H}$ and data set ${\cal X}$, we should try to
find the hypothesis or a combination of hypotheses in ${\cal H}$ that
compresses ${\cal X}$ most. The idea is applied successfully to the
problems of model selection and overfitting~\citep{grunwald05mdl}.

For a sequence of $m$ data points $\mathscr{X}$ we want to find a
model defined by a set of parameters $\boldsymbol{\theta}=[\theta_1,
  \ldots, \theta_L]$ that can efficiently maximize
$P_{\boldsymbol{\theta}}(\mathbf{x})$ for the data $\mathbf{x}$. The
description length ${\cal M}_j(\mathbf{x})$ for data $\mathbf{x}$ of
an underlying model $j$ is given by,
\begin{equation}
  \label{eq:mdl_definition}
  {\cal M}_j(\mathbf{x}) = -\log \left(
  P_{\hat{\boldsymbol{\theta}}^{(j)}}(\mathbf{x})\right) +
  \ell(\hat{\boldsymbol{\theta}}^{(j)})
\end{equation}
where
$-\log\left(P_{\hat{\boldsymbol{\theta}}^{(j)}}(\mathbf{x})\right)$
represents the maximum likelihood estimate of model $j$ which is
identical to the negative of the log-likelihood
(Section~\ref{sec:dbn}); and $\ell(\hat{\boldsymbol{\theta}}^{(j)})$
is the complexity of the model. Since the ultimate goal is to treat
the DBN as a regular DNN, the complexity of the regular DNN is defined
as the Rademacher complexity of Equation~\eqref{eq:complexity}.

Appendix~\ref{sec:algo_expansion} describes in detail the general
algorithm. In summary, the algorithm assumes there is a maximum
number of layers $T$. For every time step $t$ for a network $U_{t+1}$
we will train a new candidate layer $\mathbf{h}_{t+1}'$. In order for
the candidate layer to become part of the new network $U'_{t+1}$ it
must maximize the log-likelihood taking into account
Equation~\ref{eq:mdl_definition}.


\section{Experimental results}
\label{sec:experiments}
In this section we present the performance of architectures generated
by \textsc{EnergyNet} on a number of datasets: \texttt{mnist},
\texttt{cifar-10}, \texttt{german} and \texttt{diabetes}. Note that
convolutional neural networks are often a more natural choice for
image classification problems such as \texttt{cifar-10}
($60\mathord,000$ images evenly categorized in $10$ different
classes~\citep{Krizhevsky09learningmultiple}. However, the goal of our
experiments was a proof-of-concept showing that the structural
learning approach we propose is very competitive with traditional
approaches for finding efficient architectures.

\begin{table}[htp]
  \caption{Architecture and accuracy results for different datasets.}
  \label{tbl:accuracies}
  \centering
  \begin{tabular}{ccc|cc}
    \toprule
    & \multicolumn{2}{c}{Best architecture} & \multicolumn{2}{c}{Accuracy} \\
    \cmidrule{2-3} \cmidrule{4-5}
    Dataset & DNN & EnergyNet & DNN & EnergyNet \\
    \midrule
    \texttt{mnist} & 2048, 512, 1024 & 1525, 239 & $0.9862 \pm 0.0003$ & \bm{$0.9873 \pm 0.0002$} \\
    \texttt{cifar-10} & 256, 1024 & 101, 257 & \bm{$0.6024 \pm 0.0031$} & $0.5856 \pm 0.0011$ \\
    \texttt{german} & 16, 32 & 122 & $0.8677 \pm 0.010$ & \bm{$0.8851 \pm 0.0079$} \\
    \texttt{diabetes} & 32, 16 & 184 & $0.8433 \pm 0.011$ & \bm{$0.8593 \pm 0.0078$} \\
    \bottomrule
  \end{tabular}
\end{table}

In Table~\ref{tbl:accuracies} we compare \textsc{EnergyNet} to a
regular framework for defining DNN. The regular framework uses the
following hyperparameter ranges: learning rate in $[0.0001, 0.5]$,
batch size in $[16, 32, 64, 128]$ and ReLu activations. The search for
hidden layer architectures is performed using a combination of 30
networks with a predefined number of nodes using a Gaussian process
bandits algorithm~\citep{Snoek12bandit}.

The \textsc{EnergyNet} framework fixes the architecture with a maximum
number of layers of 10, dual norm of 2, a constant $\beta=1.01$ and
the same learning rate of 0.01 for every
dataset. Appendix~\ref{sec:algo_expansion} describes how the growth in
every layer is controlled via $\Gamma$. This is fixed to $\Gamma=0.1$
for all datasets but \texttt{mnist} which is set to $\Gamma=0.8$.
Once the architecture is fixed, a DNN is trained and tuned using the
same range of hyperparameters as in the regular framework.


\section{Conclusions}
We presented \textsc{EnergyNet}: a new framework for unsupervisedly
analyzing the architecture of artificial neural networks. Our method
optimizes for reconstruction performance, and it explicitly and
automatically addresses the trade-off between network architecture and
data modeling.
We presented experimental results showing that \textsc{EnergyNet} can
efficiently learn DNN architectures that achieve comparable results
with baseline methods.
\textsc{EnergyNet} generates DNNs with fewer number of
parameters. This is caused by the greedy construction of the model
layer by layer with finer granularity in the total number of nodes in
each layer.
Our framework is independent of the type of input features so it can
be used for other purposes as well (e.g. as a weak learner in
\textsc{AdaNet}~\citep{cortes16adanet}).

\bibliographystyle{abbrvnat}
\bibliography{paper}

\appendix

\section{Algorithm}
\label{sec:algo_expansion}
In the following section we present the details of the algorithm
introduced in Section~\ref{sec:algo}.

The algorithm runs for $T$ time steps in order to create a maximum of
$T$ layers (see Algorithm~\ref{alg:mdl}, function
$\textsc{BuildNetwork}$)).
At time $t$ there exist a neural network $U_t$ (see
Figure~\ref{fig:networks}). A candidate layer $\mathbf{h}_{t+1}'$ is
trained as a new iRBM model. This iRBM is designed to have an infinite
number of nodes. This requires that a finite number of neurons $n$ is
selected (see \textsc{TrainNewLayer}).

\begin{algorithm}[h]
  \caption{Constructs a network where $\gamma$ is the average growth
    of a layer on each training step and $\Gamma$ is a threshold. A
    layer stops growing when the number of nodes converges.}
  \label{alg:mdl}
  \begin{algorithmic}
    \Function{TrainNewLayer}{$\mathbf{v}$}
    \State $n \gets 0$
    \State $c \gets 0$
    \While{$\gamma > \Gamma$}
    \If{\Call{AddNode}{\null}}
    \State $n \gets n + 1$
    \EndIf
    \State $\gamma \gets \nicefrac{n}{c}$
    \State $c \gets c + 1$
    \EndWhile
    \EndFunction
    \Statex

    \Function{BuildNetwork}{}
    \State $U \gets \{\}$
    \For{$t < T$}
    \State $\mathbf{h}_t' \gets$ \Call{TrainNewLayer}{$\mathbf{v}$}
    \State ${\cal L} \gets$ \Call{CalculateLoglikehood}{$\mathbf{x}$}
    \State ${\cal M}_t \gets -{\cal L} + \ell(U')$
    \If{${\cal M}_t \ge {\cal M}_{t-1}$}
    \Return $U$
    \Else
    \State $U \gets \mathbf{h}_t'$
    \EndIf
    \EndFor
    \State \Return $U$
    \EndFunction
  \end{algorithmic}
\end{algorithm}

The description length of a fitted model ${\cal M}_t$ is the sum of
two parts. The first part of the description length represents the fit
of the model to the data; as the model fits better, this term
shrinks. The second part of the description length represents the
complexity of the model. This part encodes the parameters of the model
itself; it grows as the model becomes more complex.

\begin{figure}[t]
  \centering
  \begin{minipage}{.45\textwidth}
    \centering
    \scalebox{.8}{\drawnn{2}{1}{3,3}{center}}
    \caption*{Network $U_t$.}
  \end{minipage}
  \begin{minipage}{.45\textwidth}
    \centering
     \scalebox{.8}{\drawnn{2}{1}{3,3,4}{center,last_layer_rect}}
    \caption*{Network $U_{t+1}$.}
  \end{minipage}
  \caption{Network construction at time $t$. Output connections are
    marked with dotted lines because they do not exist at construction
    time.}
  \label{fig:networks}
\end{figure}

\end{document}